\documentclass{article} 
\usepackage[preprint]{colm2026_conference}

\usepackage{microtype}
\usepackage{hyperref}
\usepackage{url}
\usepackage{booktabs}
\usepackage{amsmath,amssymb,mathtools}
\usepackage{multirow}
\usepackage{lineno}
\usepackage{graphicx}
\definecolor{darkblue}{rgb}{0, 0, 0.5}
\hypersetup{colorlinks=true, citecolor=darkblue, linkcolor=darkblue, urlcolor=darkblue}

\title{Cross-Stage Coherence in Hierarchical Driving VQA: A Consistency Protocol and Parameter-Efficient Gated Context Routing}

\author{
Gautam Kumar Jain \\
Technische Hochschule Augsburg \\
TTZ Landsberg, Landsberg am Lech, Germany \\
\texttt{gautam.kumar.jain@tha.de}
\And
Carsten Markgraf \\
Technische Hochschule Augsburg \\
TTZ Landsberg, Landsberg am Lech, Germany \\
\texttt{carsten.markgraf@tha.de}
\And
Julian St\"ahler \\
Technische Hochschule Augsburg \\
TTZ Landsberg, Landsberg am Lech, Germany \\
\texttt{julian.staehler@tha.de}
}
\begin{document}

\ifcolmsubmission
\linenumbers 
\fi

\maketitle
\lhead{Preprint. Presented at COLM 2026 Workshop on Efficient Reasoning.}
\begin{abstract}

Graph Visual Question Answering (GVQA) for autonomous driving organizes reasoning into ordered stages, namely Perception, Prediction, and Planning, where planning decisions should remain consistent with the model's own perception. We present a comparative study of cross-stage context passing on DriveLM-nuScenes using two complementary mechanisms. The explicit variant evaluates three prompt-based conditioning strategies on a domain-adapted 4B VLM (Mini-InternVL2-4B-DA-DriveLM) without additional training, reducing NLI contradiction by up to $42.6\%$ and establishing a strong zero-training baseline. The implicit variant introduces gated context projectors, which extract a hidden-state vector from one stage and inject a normalized, gated projection into the next stage's input embeddings. These projectors are jointly trained with stage-specific QLoRA adapters on a general-purpose 8B VLM (InternVL3-8B-Instruct) while updating only ${\sim}0.5\%$ of parameters. The implicit variant achieves a statistically significant $34\%$ reduction in planning-stage NLI contradiction (bootstrap 95\% CIs, $p<0.05$) and increases cross-stage entailment by $50\%$, evaluated with a multilingual NLI classifier to account for mixed-language outputs. Planning language quality also improves (CIDEr $+30.3\%$), but lexical overlap and structural consistency degrade due to the absence of driving-domain pretraining. Since the two variants use different base models, we present them as complementary case studies: explicit context passing provides a strong training-free baseline for surface consistency, while implicit gated projection delivers significant planning-stage semantic gains, suggesting domain adaptation as a plausible next ingredient for full-spectrum improvement.

\end{abstract}

\section{Introduction}
\label{sec:intro}
Autonomous driving research is shifting toward systems that produce interpretable, multi-step reasoning before committing to an action~\cite{drivelm,driveVLM,lmdrive}. Vision-language models (VLMs), pretrained on large multimodal corpora, carry priors about object behavior, traffic rules, and causal relationships that purely geometric planners lack~\cite{CosmosReason1,AlphaDrive,SimLingo}.

Graph Visual Question Answering (Graph VQA), introduced by DriveLM~\cite{drivelm}, organizes driving-based reasoning into three ordered stages: Perception ($\mathcal{P}_\text{erc}$), Prediction ($\mathcal{P}_\text{red}$), and Planning ($\mathcal{P}_\text{lan}$). The ordering reflects real dependencies: a prediction that ignores perception is likely incoherent; a plan that ignores prediction is potentially unsafe. GVQA therefore serves as a structured diagnostic where cross-stage information flow can be measured directly.

A model that answers each stage independently risks outputs that are locally fluent but globally contradictory~\cite{drivebench}. End-to-end architectures~\cite{DriveTransformer,HiPAD} pass structured feature tokens between tasks, but these encode physical scene elements, not compressed semantic outputs. Chain-of-thought methods~\cite{DriveLMMo1,DriveCoT,OpenEMMA} rely on the autoregressive context window, which is not trainable as a routing mechanism. Stage-specific adapters~\cite{DriveAdapter} train each adapter independently, making cross-stage coherence accidental. What is missing is a trainable mechanism that propagates a compact semantic state from one reasoning stage to the next.

We approach this through two complementary forms of cross-stage context passing (Figure~\ref{fig:context_passing}). In the \textit{explicit} form, context passes between stages as generated text through one of three prompt-based conditions; no training is involved. In the \textit{implicit} form, a compact vector from the VLM's last-layer hidden states is projected and gated into the next stage's input embeddings through a learned linear module, within a single inference pipeline.

The explicit variant uses Mini-InternVL2-4B-DA-DriveLM~\cite{miniinternv-dal, miniinternvl}, a domain-adapted model; the implicit variant uses InternVL3-8B-Instruct~\cite{internvl3_8b_instruct,internvl3}, a general-purpose VLM without driving pretraining. Results across variants are not directly comparable; we report them as complementary case studies. Our contributions are:
\begin{figure}[t]
\centering
\includegraphics[width=\linewidth]{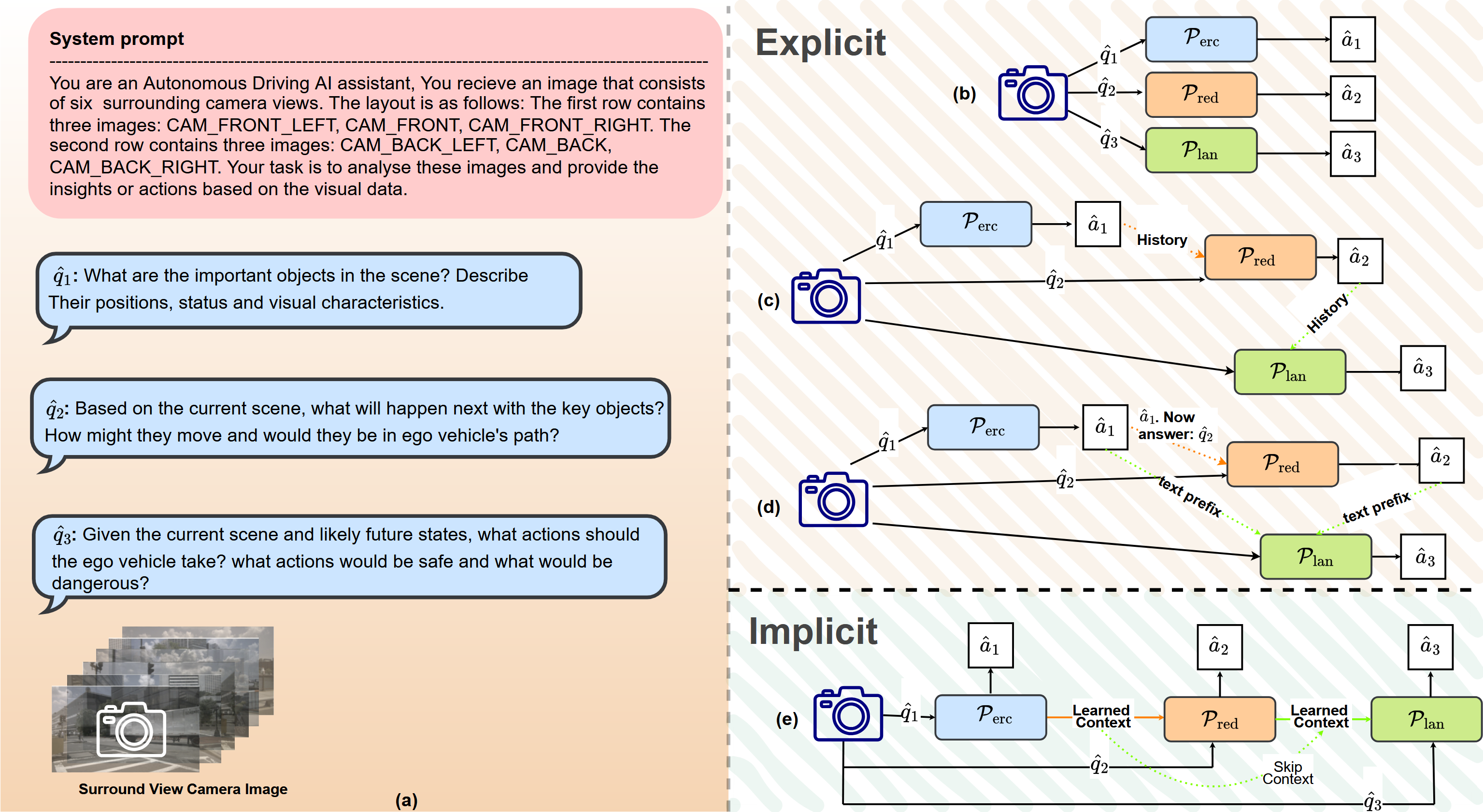}
\caption{%
  \textbf{Context passing mechanisms for hierarchical driving VQA.}
  (a)~Shared setup: six surround-view cameras and three stage-specific questions.
  (b)~Flat: each stage processes image and question independently.
  (c)~History-chain: prior answers flow as conversational history.
  (d)~Injection-chain: prior answers are prepended as structured text prefixes.
  (e)~Proposed framework: learned context vectors (\texttt{perception\_to\_prediction}, \texttt{prediction\_to\_planning}) are projected between stages via gated linear modules; an optional skip routes perception context directly to planning.}
\label{fig:context_passing}
\end{figure}

\begin{enumerate}
\item \textbf{Explicit context-passing study} using structured prompt injection on a domain-adapted 4B VLM, showing that hierarchical conditioning reduces NLI contradiction by up to $42.6\%$ with no training.

\item \textbf{Modular implicit reasoning framework} with three stage-specific QLoRA adapters on InternVL3-8B, connected by gated context projectors that route compressed semantic state between stages, training only ${\sim}0.5\%$ of parameters.

\item \textbf{NLI-based cross-stage consistency protocol.} An evaluation framework for hierarchical driving VQA that combines lexical overlap, rule-based structural consistency, and NLI-based contradiction scores across adjacent stage pairs, a property absent from existing driving VQA benchmarks.
\item \textbf{Domain adaptation as a missing ingredient.} A diagnostic finding: implicit gated projection significantly improves planning-stage NLI coherence (bootstrap 95\% CIs, $p<0.05$) regardless of base model, but surface-level consistency (lexical overlap, structural extraction) requires driving-domain pretraining. This identifies a plausible prerequisite for future work applying implicit context routing to driving VQA.
\end{enumerate}

\section{Related Work}
\label{sec:relwork}
\subsection{Vision-Language Models for Autonomous Driving}
VLMs have been applied to driving as reasoning components within planning pipelines~\cite{driveVLM,lmdrive,AlphaDrive,OpenEMMA} and as end-to-end systems~\cite{SimLingo,opendrivevla,dima,vlad}. These works target closed-loop control or trajectory planning. Our work operates in the GVQA diagnostic setting, using it as a testbed for cross-stage state passing rather than closed-loop control.

\subsection{Hierarchical and Multi-Stage Reasoning}
In end-to-end planning, DriveTransformer~\cite{DriveTransformer}, HiP-AD~\cite{HiPAD}, CogAD~\cite{CogAD}, and PARA-Drive~\cite{paradrive} propagate structured feature tokens tied to physical entities between task heads, not compressed semantic states from natural-language answers.

In VQA, DriveLM~\cite{drivelm} links perception--prediction--planning QA pairs via a dependency graph; DriveLMM-o1~\cite{DriveLMMo1} and DriveCoT~\cite{DriveCoT} generate step-by-step reasoning chains but rely on the autoregressive context window rather than a trainable routing mechanism. Hierarchical QA for driving~\cite{HierarchicalQA} uses a hand-crafted question tree with fixed templates.

Hint-AD~\cite{HintAD}, VERDI~\cite{VERDI}, and ALN-P3~\cite{alnp3} improve representations via language grounding across the perception-prediction-planning stack but do not route a learned semantic state between VQA stages at inference time.

\paragraph{PEFT and benchmarks.}

LoRA~\cite{LoRA} and QLoRA~\cite{QLoRA} enable parameter-efficient adaptation; DriveAdapter~\cite{DriveAdapter} applies feature-alignment adapters across module boundaries but trains each stage independently; MMA~\cite{mma} extends adapter placement to both branches. The DriveLM benchmark~\cite{drivelm} evaluates per-stage language quality (top entry: Precise Drive~\cite{Huang2024PreciseDW}, composite 0.6064); DriveBench~\cite{drivebench} adds robustness evaluation. A property absent from these benchmarks is \textit{cross-stage consistency}: whether a model's Planning answer is semantically coherent with its own Perception answer.
\paragraph{Cross-stage capability alignment.}
Drive-KD~\cite{Lian2026DriveKDMD} decomposes autonomous-driving VLMs into a perception-reasoning-planning triad and finds that these capabilities conflict under joint optimization; it introduces asymmetric gradient projection to mitigate the resulting cross-capability gradient conflicts at training time.
Drive-P2D~\cite{Tang2026DriveP2DAP} provides direct empirical evidence: a progressive perception-to-decision benchmark across Object, Scene, and Decision levels reveals weak alignment between perception and decision-making performance (Pearson $\rho \in [-0.2, 0.2]$), framed as an open challenge for driving VLMs. Both works demonstrate that perception, reasoning, and planning are weakly coupled and require explicit coordination. Our gated context projectors address the same coupling problem at a complementary \emph{inference-time} axis: lightweight learned routing across adapter-separated stages, applicable on top of already-trained adapters without modifying training-time objectives.

\paragraph{Remaining gap.}
Recent latent-reasoning work (COCONUT~\cite{coconut}; CODI~\cite{codi}) routes hidden states forward within a single homogeneous chain, improving reasoning efficiency. DriveAdapter~\cite{DriveAdapter} trains stage-specific adapters independently, making cross-stage coherence accidental. Neither approach evaluates whether hidden-state routing reduces semantic contradiction between discrete, adapter-separated reasoning stages with different semantic objectives.

The contribution of this paper is not the gated projection primitive itself, which is a known building block, but two novel elements: (i)~the \emph{cross-stage adapter-separated training regime} (frozen upstream adapter, jointly trained downstream adapter and projector), which has not been evaluated for driving VQA; and (ii)~an \emph{NLI-based cross-stage consistency protocol}, absent from DriveLM~\cite{drivelm} and DriveBench~\cite{drivebench}. Together they enable a controlled within-model study of whether learned hidden-state routing reduces planning-stage semantic contradiction.

\section{Problem Formulation}
\label{sec:formulation}

\subsection{GVQA as a Structured Reasoning Testbed}

Let $\mathbf{v} \in \mathcal{V}$ denote a multi-camera observation of a driving scene (six synchronized RGB frames from the nuScenes sensor suite~\cite{nuscenes}).
The reasoning stages are $\mathcal{S} = \{s_1, s_2, s_3\} = \{\mathcal{P}_\text{erc}, \mathcal{P}_\text{red}, \mathcal{P}_\text{lan}\}$, with a causal dependency: planning should follow from prediction, which should follow from perception.

For each stage $s_k$ there is a set of $M_k$ question-answer pairs $\mathcal{Q}_k = \{(q_{k,j},\, a^*_{k,j})\}_{j=1}^{M_k}$.
Let $f_\theta$ denote a VLM parameterized by $\theta$.
The independent (flat) baseline $\hat{a}_{k,j}^{\,\text{flat}} = f_\theta(\mathbf{v},\, q_{k,j})$ provides no mechanism for cross-stage coherence. The goal is to define hierarchically conditioned policies $\hat{a}_{k,j}^{\,\text{hier}}$ that condition on the output of stage $s_{k-1}$ when answering stage $s_k$, and to measure whether this reduces cross-stage contradictions.

\subsection{Explicit Reasoning: Prompt-Based Context Passing}
\label{ssec:form_explicit}

In the explicit variant, context passes between stages as generated text, without any trainable component. We evaluate three conditions on Mini-InternVL2-4B-DA-DriveLM.

\paragraph{Condition 1 (Flat).}
Each stage receives only the visual input and its own question:
\[
\hat{a}_{k}^{\,\mathrm{flat}} = f_\theta(\mathbf{v},\, q_k).
\]

\paragraph{Condition 2 (History-chain).}
All three stages are posed within a single multi-turn conversation. Stage~1 is answered first; the model's autoregressive memory carries its response into subsequent turns:
\[
\hat{a}_{k}^{\,\mathrm{hist}}
= f_\theta\!\bigl(\mathbf{v},\; q_{k} \;\big|\; \mathrm{history}(\hat{a}_{1}, \ldots, \hat{a}_{k-1})\bigr),
\]
where $\mathrm{history}(\cdot)$ denotes the conversation state maintained by the VLM's chat interface.

\paragraph{Condition 3 (Injection-chain).}
Each stage is a fresh single-turn call, but prior-stage answers are prepended as a structured text prefix:
$\hat{a}_{k}^{\,\mathrm{inj}} = f_\theta\!\bigl(\mathbf{v},\; \pi_k(\hat{a}_1, \ldots, \hat{a}_{k-1},\, q_k)\bigr)$,
where $\pi_k$ concatenates prior answers as labeled fields, e.g.\ $\pi_3 = \texttt{"Perception: } \hat{a}_1 \texttt{. Prediction: } \hat{a}_2 \texttt{. Now answer: } q_3\texttt{"}$.

\begin{figure}[t]
\centering
\includegraphics[width=\linewidth]{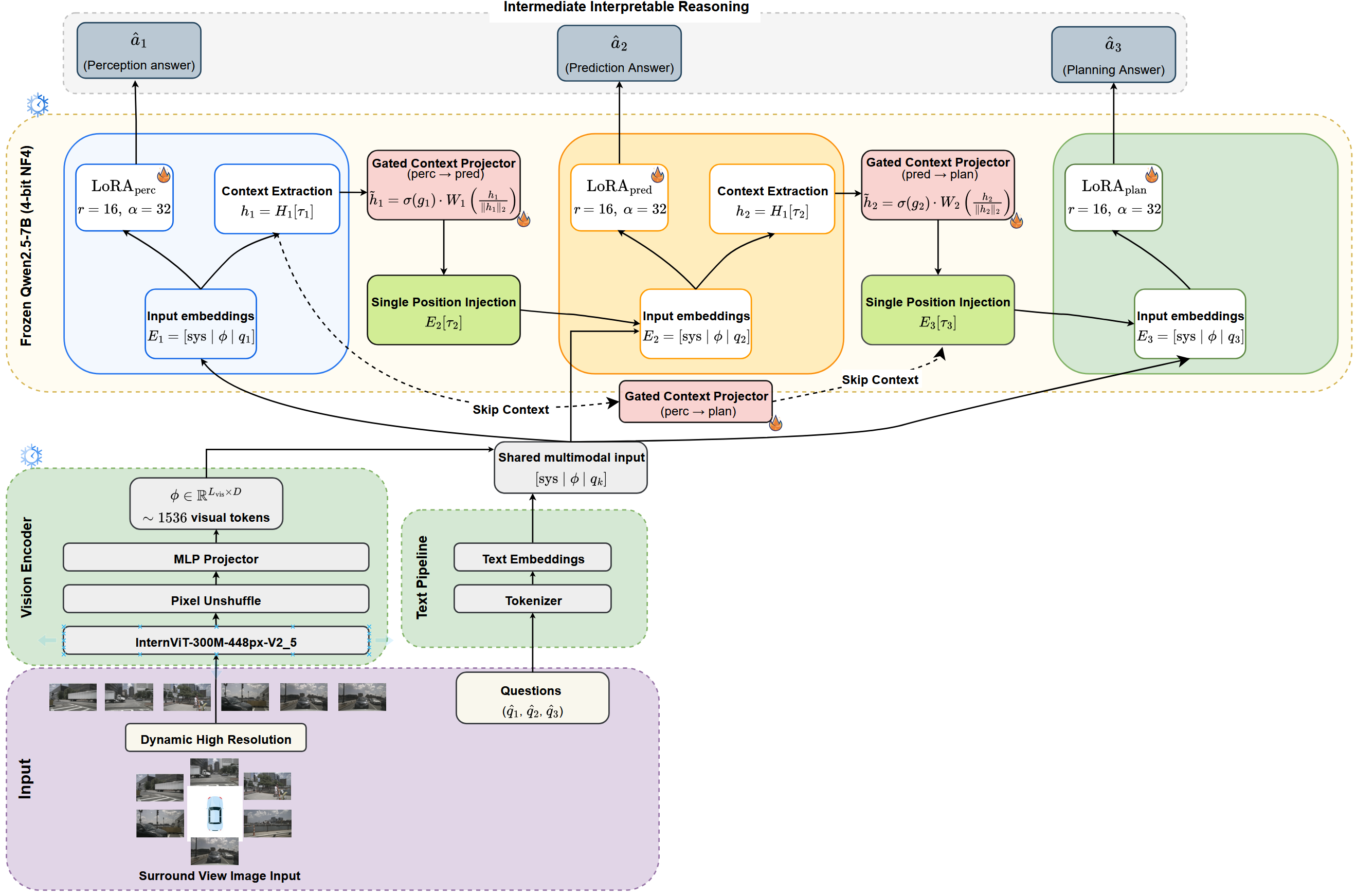}
\caption{%
  \textbf{Architecture of the implicit variant.}
  The frozen InternVL3-8B-Instruct backbone carries three stage-specific LoRA adapters.
  After each stage, a gated context projector extracts the hidden state at the final prompt token and injects a normalized, gated projection into the input embeddings of the next stage.
  The full three-stage chain executes in a single inference pipeline.
}
\label{fig:architecture}
\end{figure}

\subsection{Implicit Reasoning: Learned Gated Context Projection}
\label{ssec:form_implicit}

The implicit variant replaces text-level context passing with a trainable mechanism operating in the VLM's hidden-state space ($D = 3584$ for InternVL3-8B). We describe three components.

\paragraph{Semantic context extraction.}
Let $\mathbf{H}_k \in \mathbb{R}^{L \times D}$ denote the last-layer hidden states after processing stage $s_k$, where $L$ is the sequence length. The context vector $h_k \in \mathbb{R}^D$ is the hidden state at position $\tau_k$:
\[
h_k = \mathbf{H}_k[\tau_k],
\]
where $\tau_k$ is the last non-visual, non-padding prompt token. In a causal transformer, this position attends to the entire preceding context and encodes a compressed representation of the model's scene understanding conditioned on the question.

\paragraph{Gated projection.}
The context $h_k$ is transformed into a conditioning signal for stage $s_{k+1}$:
\[
\tilde{h}_k = \sigma(g_k) \cdot W_k \!\left(\frac{h_k}{\|h_k\|_2 + \varepsilon}\right), \qquad \tilde{h}_k \in \mathbb{R}^D,
\]
where $W_k \in \mathbb{R}^{D \times D}$ is a learnable weight matrix, $g_k \in \mathbb{R}$ is a learnable scalar gate, $\sigma(\cdot)$ is the sigmoid function, and $\varepsilon = 10^{-6}$. $L_2$-normalization is necessary because raw hidden-state norms (${\sim}100$--$200$) exceed token embedding norms (${\sim}4$); without it, the projected vector causes generation collapse. The gate starts near zero ($g_k = -3.5$, $\sigma(g_k) \approx 0.029$) and learns to open during training, providing a curriculum from independent to context-conditioned inference.

\paragraph{Single-position injection.}
The gated context $\tilde{h}_k$ is added to the input embedding at position $\tau_{k+1}$: $\mathbf{E}_{k+1}[\tau_{k+1}] \mathrel{+}= \tilde{h}_k$,
where $\mathbf{E}_{k+1} \in \mathbb{R}^{L \times D}$. Single-position injection avoids a distribution shift across the full sequence and keeps the injection site identical between training and inference.

\section{Experimental Setup}
\label{sec:experiments}

\subsection{Models and Dataset}
\paragraph{Explicit variant.}
Mini-InternVL2-4B-DA-DriveLM, a 4B-parameter VLM domain-adapted on DriveLM-nuScenes with a coordinate-aware loss. No additional training; all three prompt conditions are evaluated inference-only.

\paragraph{Implicit variant.}
InternVL3-8B-Instruct, a general-purpose 8B-parameter VLM with no driving-domain pretraining, loaded with 4-bit NF4 quantization (double quantization) and fine-tuned via QLoRA~\cite{QLoRA} with stage-specific LoRA adapters. The vision encoder and all non-adapter parameters remain frozen.

\paragraph{Dataset.}
Both variants use DriveLM-nuScenes~\cite{drivelm,nuscenes} (v1.1, ${\sim}3{,}200$ keyframes, 80/20 split): six camera views stitched into a $1344 \times 896$ image with stage-specific QA pairs. Consistency analysis uses three fixed open-ended questions across 796 validation scenes (Figure~\ref{fig:context_passing}).

\subsection{Training Protocol}
\paragraph{Flat adapter baselines.}
Three independent LoRA adapters (one per stage) trained separately on their respective QA pairs with the full backbone frozen.

\paragraph{Sequential training with context projection.}
Training follows the pipeline $\mathcal{P}_\mathrm{erc} \to \mathcal{P}_\mathrm{red} \to \mathcal{P}_\mathrm{lan}$ in two phases. In Phase~1 ($\mathcal{P}_\mathrm{erc}^{*} \to \mathcal{P}_\mathrm{red}$), the trained Perception adapter is frozen ($^{*}$) and a new Prediction adapter is trained jointly with projector $W_1$ and gate $g_1$. In Phase~2 ($\mathcal{P}_\mathrm{erc}^{*} \to \mathcal{P}_\mathrm{red}^{*} \to \mathcal{P}_\mathrm{lan}$), both upstream adapters are frozen and a Planning adapter is trained with projector $W_2$ and gate $g_2$. An optional skip connection adds projector $W_{1 \to 3}$ routing Perception context directly into Planning. In Phase~2, $W_2$ and $g_2$ can optionally be initialized from the trained $W_1$ and $g_1$ of Phase~1, providing a starting point that has already learned to route hidden-state context.


\paragraph{Why different base models.}
We first attempted to attach sequential stage-specific LoRA adapters to Mini-InternVL2-4B-DA-DriveLM, but this produced progressive quality degradation: the DA model is jointly optimized across all three stages with a coordinate-aware loss, and stacking per-stage adapters on that backbone disturbs the cross-stage representations it has already encoded.
The explicit variant is unaffected because prompt-level context passing is orthogonal to the training regime; the implicit variant therefore uses InternVL3-8B-Instruct with freshly initialized adapters, providing a clean base for sequential training.
A unified-backbone comparison would require training a domain-adapted InternVL3-8B with a per-stage adapter protocol from scratch, which we identify as the natural next step (Section~\ref{sec:conclusion}).

\paragraph{Hyperparameters.}

All LoRA adapters: rank $r = 16$, $\alpha = 32$, dropout $0.05$, applied to all attention and MLP projection layers. Backbone: BF16 with 4-bit NF4 double quantization. Optimizer: AdamW, base LR $1.5 \times 10^{-5}$, weight decay $0.05$, cosine schedule with $10\%$ warmup. Context projectors receive $3.0\times$ the base LR; gates receive $25.0\times$. Projector weights initialized at scale $0.01$; gate scalars at $g_k = -3.5$. Per-adapter deterministic seeds ensure identical initialization between flat and sequential variants. Total trainable parameters: ${\sim}0.5\%$ of the 8B backbone.

\subsection{Evaluation Metrics}
\paragraph{Language quality.}
Per-stage answers scored against DriveLM ground truth using BLEU-1~\cite{Bleu}, ROUGE-L~\cite{rouge}, and CIDEr~\cite{CIDEr}.

\paragraph{Cross-stage consistency.}
Three complementary metrics computed over the fixed open-ended questions across all validation scenes.
\textit{Lexical overlap:} For each ordered stage pair $(s_k, s_{k+1})$, key terms are extracted (tokens $>3$ characters after stop-word removal) and source-term recall in the target is computed:
\[
\mathrm{LexOverlap}(s_k, s_{k+1}) = \frac{|\mathrm{terms}(\hat{a}_k) \cap \mathrm{terms}(\hat{a}_{k+1})|}{|\mathrm{terms}(\hat{a}_k)|}.
\]
\textit{Structural consistency:} A rule-based system extracts driving-domain attributes (traffic lights, ego actions, pedestrian status) and checks for logical contradictions: $\mathrm{StructConsist} = 1 - \text{contradictions}/\text{rule checks}$, counted only when both answers contain extractable attributes.

This metric has a known blind spot: it can reward an internally contradictory hedged plan (e.g., listing both ``brake to a stop'' and ``keep going at the same speed'') over a decisive single-action plan that omits a safety-relevant keyword. This failure mode is illustrated qualitatively in Appendix~\ref{sec:appendix} (Scene 360) and is why NLI is our load-bearing semantic metric.

\textit{NLI-based consistency:} A multilingual NLI classifier (mDeBERTa-v3-base-XNLI, 100 languages~\cite{Laurer2023BuildingEU}) scores each adjacent answer pair for entailment, neutrality, and contradiction across $\mathcal{P}_\mathrm{erc} \to \mathcal{P}_\mathrm{red}$ and $(\mathcal{P}_\mathrm{erc} + \mathcal{P}_\mathrm{red}) \to \mathcal{P}_\mathrm{lan}$. The primary metric is average contradiction probability (lower is better). A multilingual classifier is essential because the implicit variant produces mixed-language outputs (Section~\ref{sec:diagnostic}); a monolingual classifier would conflate language shift with semantic failure.

\section{Results}
\label{sec:results}

\subsection{Context Projector Dynamics}

\begin{figure}[t]
\begin{center}
\includegraphics[width=0.75\linewidth]{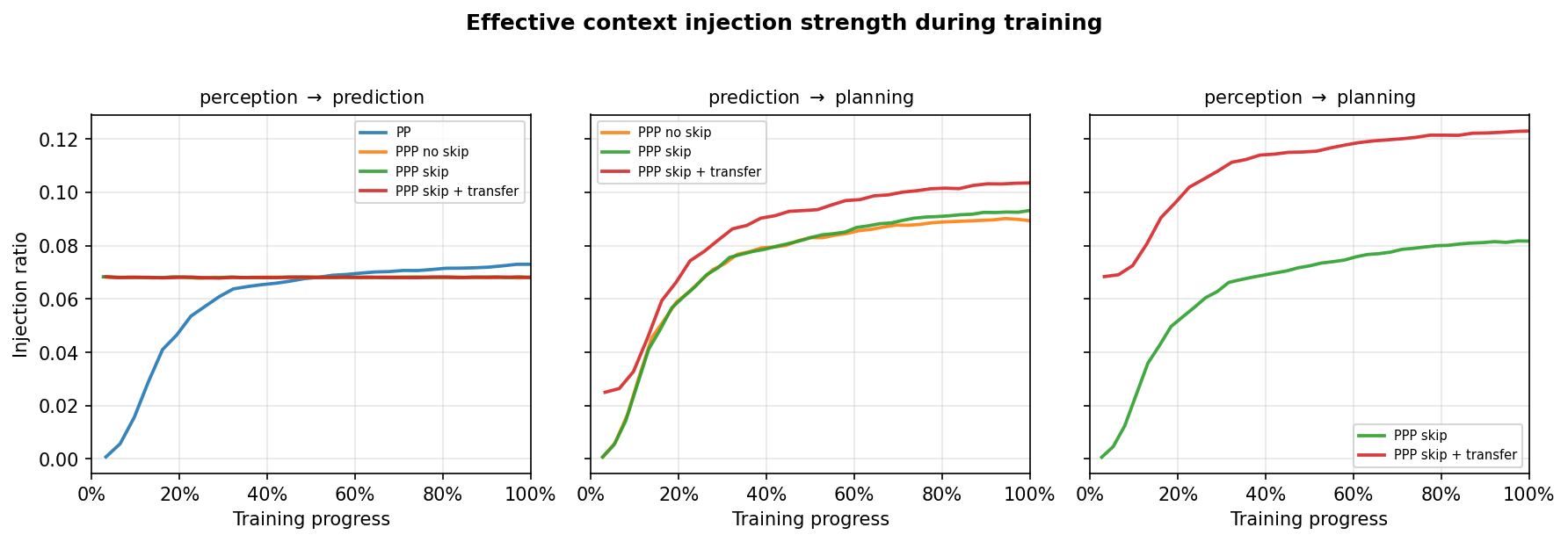}
\vspace{0.3em}
\includegraphics[width=0.75\linewidth]{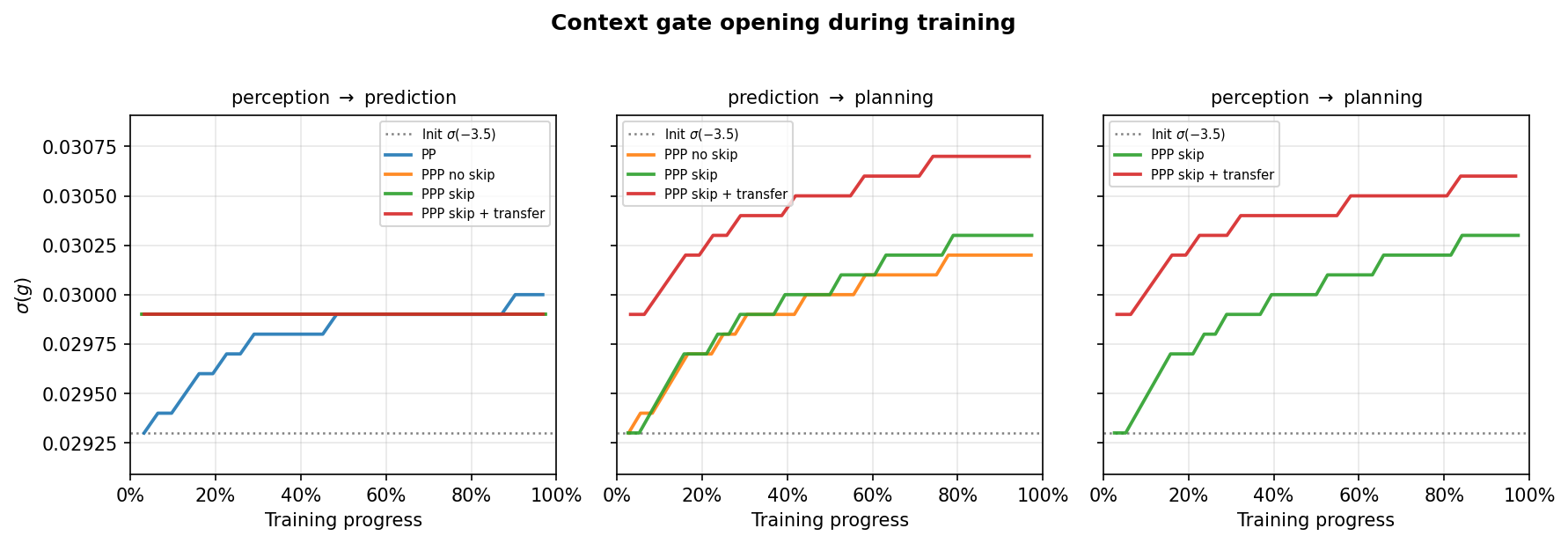}
\end{center}
\caption{Effective context injection strength (top) and gate opening $\sigma(g_k)$ (bottom) during training, for three projector transitions.}
\label{fig:projector_dynamics_combined}
\end{figure}

Figure~\ref{fig:projector_dynamics_combined} shows the effective injection ratio $\|\tilde{h}_k\|_2 / \|\mathbf{E}_{k+1}[\tau_{k+1}]\|_2$ and gate opening $\sigma(g_k)$ over training. The perception$\to$prediction injection ratio rises to ${\sim}0.07$ in Phase~1, while the gate increases only marginally ($\sigma(g_k) \approx 0.029 \to 0.030$). For prediction$\to$planning and skip projectors, weight transfer eliminates the cold-start period and yields higher final injection ratios (${\sim}0.09$--$0.12$). The near-constant gate combined with a growing injection ratio reveals that $W_k$ carries the primary conditioning signal, while $g_k$ functions as a \emph{protective} stability margin rather than a routing mechanism: initialized near-closed ($\sigma(-3.5) \approx 0.029$), it prevents injection collapse during early training before the projector weights (initialized at scale 0.01, given $3\times$ base LR) have converged.

\subsection{Per-Stage Language Quality}
Table~\ref{tab:language} reports language quality for flat and sequential variants.

\begin{table}[t]
\centering
\caption{Per-stage language quality on DriveLM-nuScenes validation (InternVL3-8B-Instruct+QLoRA). Flat adapters are trained independently. Sequential variants add learned context projectors between frozen upstream and trainable downstream adapters. $\mathcal{P}_\mathrm{erc}^{*} \to \mathcal{P}_\mathrm{red}$ evaluates prediction; PPP variants evaluate planning.}
\label{tab:language}
\resizebox{\linewidth}{!}{
\begin{tabular}{lccccccc}
\toprule
 & \multicolumn{3}{c}{\textbf{Flat Adapters}} & \textbf{Seq.} & \multicolumn{3}{c}{\textbf{Seq.\ PPP}} \\
\cmidrule(lr){2-4} \cmidrule(lr){5-5} \cmidrule(lr){6-8}
\textbf{Metric} & Perc & Pred & Plan & $\mathcal{P}^{*}\!\to\!\mathcal{P}_\mathrm{red}$ & No skip & Skip & Skip+Transf. \\
\midrule
BLEU-1$\uparrow$    & 23.8 & 36.5 & 30.1 & 33.0 & 31.5 & 32.9 & \textbf{34.1} \\
ROUGE-L$\uparrow$   & 19.0 & 33.2 & 21.6 & \textbf{39.2} & 22.9 & 24.1 & 25.3 \\
CIDEr$\uparrow$     & 50.2 &  65.1 & 51.5 & \textbf{91.4} & 55.7 & 61.6 & 67.1 \\
\bottomrule
\end{tabular}
}
\end{table}

\paragraph{Sequential training improves downstream generation.}
The sequential PPP variants outperform the flat planning baseline, with monotonic gains from no-skip to skip to skip+transfer. The strongest planning result is PPP skip+transfer: BLEU-1 34.1 (vs.\ 30.1 flat, $+13.3\%$), ROUGE-L 25.3 (vs.\ 21.6, $+17.1\%$), and CIDEr 67.1 (vs.\ 51.5, $+30.3\%$). The two-stage $\mathcal{P}_\mathrm{erc}^{*} \to \mathcal{P}_\mathrm{red}$ variant achieves the highest ROUGE-L (39.2) and CIDEr (91.4), reflecting substantial benefit from Perception context at the Prediction stage.

\subsection{Cross-Stage Consistency Analysis}
\label{sec:consistency}
\begin{table*}[t]
\centering
\caption{Cross-stage consistency on DriveLM-nuScenes. Explicit: Mini-InternVL2-4B-DA (domain-adapted, inference-only); Implicit: InternVL3-8B+QLoRA (general-purpose). NLI uses a multilingual classifier (mDeBERTa-v3-base-XNLI, 100 languages). $\downarrow$/$\uparrow$: lower/higher is better. NLI columns average both stage transitions. Bootstrap 95\% CIs confirm planning-stage NLI improvements are statistically significant (Section~\ref{sec:consistency}).}
\label{tab:consistency}
\resizebox{\linewidth}{!}{
\begin{tabular}{llcccc}
\toprule
\textbf{Variant} & \textbf{Condition} & \textbf{Lex.\ Overlap}$\uparrow$ & \textbf{Struct.\ Consist.}$\uparrow$ & \textbf{NLI-contra}$\downarrow$ & \textbf{NLI-entail}$\uparrow$ \\
\midrule
\multirow{3}{*}{\shortstack[l]{Explicit\\(4B-DA)}}
 & Flat                & 0.123 & 0.624 & 0.461 & 0.071 \\
 & History-chain       & \textbf{0.390} & \textbf{0.675} & \textbf{0.264} & \textbf{0.134} \\
 & Injection-chain     & 0.265 & 0.586 & 0.355 & 0.133 \\
\midrule
\multirow{4}{*}{\shortstack[l]{Implicit\\(8B-QLoRA)}}
 & Flat adapters       & 0.102 & 0.684 & 0.351 & 0.042 \\
 & PPP no skip         & 0.064 & 0.312 & 0.334 & 0.061 \\
 & PPP skip            & 0.054 & 0.188 & 0.335 & 0.062 \\
 & PPP skip+transfer   & 0.067 & 0.359 & \textbf{0.333} & \textbf{0.063} \\
\bottomrule
\end{tabular}
}
\end{table*}
Table~\ref{tab:consistency} reports cross-stage consistency for both variants, evaluated on 796 validation scenes.
\paragraph{Explicit variant.}Both hierarchical conditions improve consistency. History-chaining achieves the strongest results: NLI contradiction drops by $42.6\%$ ($0.461 \to 0.264$), lexical overlap triples, and entailment nearly doubles. Injection-chaining also reduces NLI contradiction ($-22.9\%$) with stable answer length. Remaining structural contradictions are dominated by ``pedestrian crossing $\to$ accelerate/maintain'' (38 of 56 flat contradictions), reflecting model bias rather than context-passing failure.

\paragraph{Implicit variant: NLI improvements.}The learned projectors concentrate their semantic effect at the planning stage: $(\mathcal{P}_\mathrm{erc} + \mathcal{P}_\mathrm{red}) \to \mathcal{P}_\mathrm{lan}$ contradiction drops by $34.4\%$ ($0.340 \to 0.223$; Table~\ref{tab:consistency} averages both transitions: $0.351 \to 0.333$), while cross-stage entailment increases by $50\%$ ($0.042 \to 0.063$). Planning is the safety-critical stage where contradictions matter most (e.g.\ perceiving pedestrians yet planning to accelerate). The average contradiction reduction ($0.351 \to 0.333$) is modest and not statistically significant, indicating that the projectors selectively improve the planning transition rather than uniformly reducing contradiction throughout the chain.

\paragraph{Implicit variant: lexical and structural degradation.}
Lexical overlap and structural consistency decrease relative to the flat baseline. The diagnostic in Table~\ref{tab:answer_length} explains this pattern.

\paragraph{Statistical significance.}
Bootstrap CIs (10{,}000 resamples, 95\% level) on 796 validation scenes confirm the planning-stage result: $(\mathcal{P}_\mathrm{erc}+\mathcal{P}_\mathrm{red}) \to \mathcal{P}_\mathrm{lan}$ contradiction flat $0.340\,[0.315,\,0.365]$ vs.\ no-skip $0.223\,[0.203,\,0.243]$ (non-overlapping, $p < 0.05$). Average entailment also improves significantly: $0.042\,[0.036,\,0.048]$ vs.\ $0.061\,[0.053,\,0.069]$. The average contradiction reduction ($0.351 \to 0.333$) is not statistically significant, as the planning-stage gain is diluted when averaged across both stage transitions. Lexical and structural degradations are also significant (non-overlapping CIs), consistent with Section~\ref{sec:diagnostic}.

\subsection{Diagnostic: Answer Length and Language Distribution}
\label{sec:diagnostic}
\begin{table}[t]
\centering
\caption{Average answer length (tokens) and language distribution for the implicit variant. Sequential training shortens Prediction outputs and introduces mixed-language generation.}
\label{tab:answer_length}
\begin{tabular}{lcccc}
\toprule
\textbf{Stage} & \textbf{Flat} & \textbf{PPP no skip} & \textbf{PPP skip} & \textbf{PPP skip+transf.} \\
\midrule
Perception   & 52.3 & 48.7 & 48.7 & 48.7 \\
Prediction   & 26.3 &  8.3 &  8.3 &  8.3 \\
Planning     & 21.6 & 20.0 & 19.5 & 20.4 \\
\bottomrule
\end{tabular}
\end{table}

Sequential variants produce shorter Prediction answers (8.3 vs.\ 26.3 tokens) and mixed-language outputs (38\% Chinese at Prediction), both tracing to the lack of driving-domain pretraining. Shorter answers lower lexical overlap mechanically; mixed-language outputs fail English-oriented rule extractors, inflating structural contradictions. Because we employ a multilingual NLI classifier (mDeBERTa-v3-base-XNLI, 100 languages), NLI scores are not confounded by the language shift: the classifier evaluates semantic coherence regardless of output language. The planning-stage NLI improvement (Section~\ref{sec:consistency}) therefore reflects genuine semantic alignment, not an evaluation artifact.

\subsection{Qualitative Results}
\label{sec:qualitative}
\begin{figure}[t]
\centering
\includegraphics[width=0.75\linewidth]{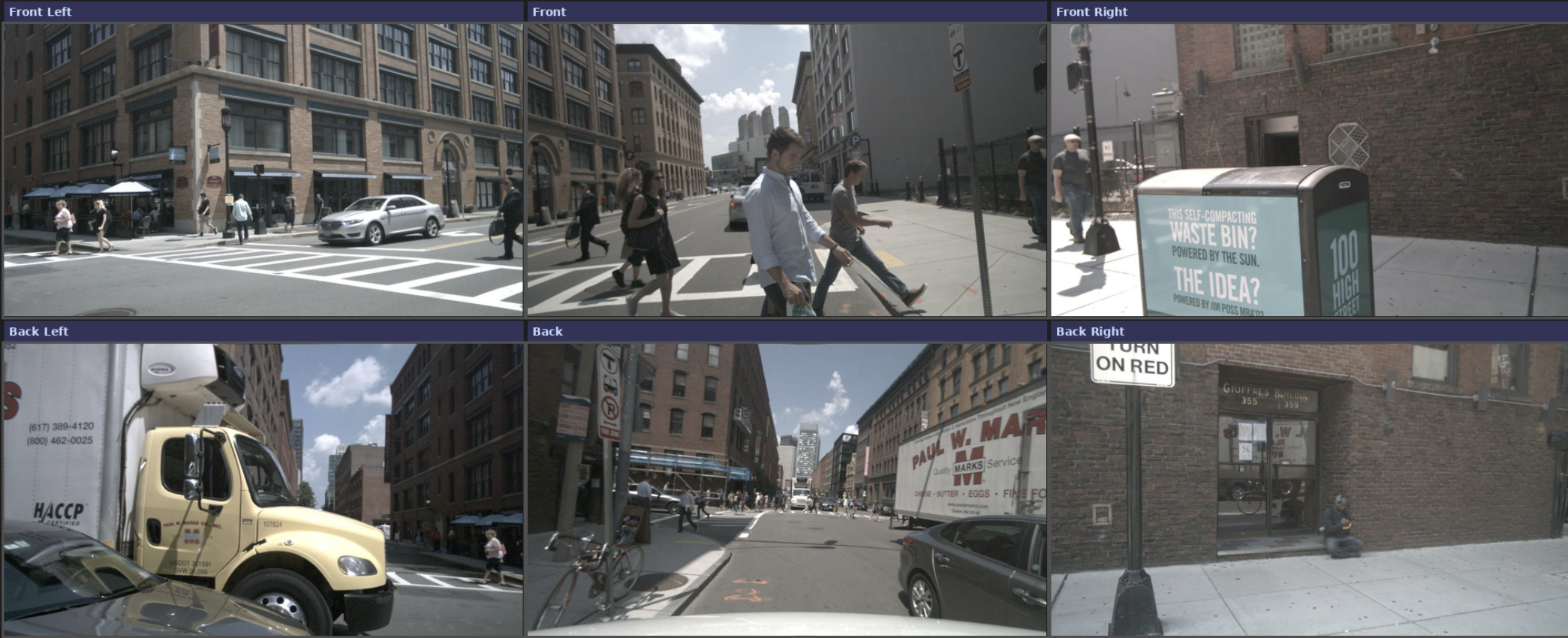}

\scriptsize
\setlength{\tabcolsep}{3pt}
\begin{tabular}{@{}c@{~}p{0.9\columnwidth}@{~}c@{}}
\toprule
& \textbf{Stage Outputs} & \textbf{C} \\
\midrule
\multicolumn{3}{@{}l}{\textit{\textbf{Explicit variant (Mini-InternVL2-4B-DA)}}} \\[2pt]
Flat &
  \textbf{P:} Many pedestrians and a car [\ldots] carrying a backpack and \underline{crossing the road}. [\ldots] \newline
  \textbf{Pr:} Keep going straight, change to the left lane, slightly offset to the left. \newline
  \textbf{Pl:} The action is to \underline{keep going at the same speed}, and the action is to slightly offset to the right.
& \textbf{1} \\[3pt]
Inj. &
  \textbf{P:} A man wearing a white shirt to the front [\ldots] a silver sedan [\ldots] and a pedestrian to the back left [\ldots] \newline
  \textbf{Pr:} The man wearing a white shirt will turn right, the silver sedan will turn right [\ldots] \newline
  \textbf{Pl:} The action is to keep going at the same speed, the action is to turn right, and the action is to \underline{stop}.
& 0 \\
\midrule
\multicolumn{3}{@{}l}{\textit{\textbf{Implicit variant (InternVL3-8B + QLoRA)}}} \\[2pt]
Flat &
  \textbf{P:} [\ldots] A silver car on the left side. \underline{Pedestrians} walking on the sidewalk. [\ldots] \newline
  \textbf{Pr:} Pedestrian 1: Will cross the road. No, they will not be in the ego vehicle's path. \newline
  \textbf{Pl:} \underline{Keep going at the same speed. Accelerate and go ahead.}
& \textbf{1} \\[3pt]
Seq. &
  \textbf{P:} Brick buildings; \underline{pedestrians} on sidewalk near building entrance; silver car parked on roadside. \newline
  \textbf{Pr:} They may enter the ego path and continue crossing. \newline
  \textbf{Pl:} The action that the ego vehicle should take is to slow down. The action that would be dangerous is to \underline{brake suddenly}.
& 0 \\
\bottomrule
\end{tabular}
\caption{Success case for Scene 162. Stage outputs (\textbf{P}/\textbf{Pr}/\textbf{Pl}) are verbatim (abbreviated). \textbf{C}: structural contradiction count. \underline{Underlines} mark contradiction triggers/resolutions. Both flat baselines perceive pedestrians but plan to maintain speed; hierarchical variants resolve the contradiction.}
\label{fig:qualitative_combined}
\end{figure}

Figure~\ref{fig:qualitative_combined} shows a representative success case. Both flat baselines perceive pedestrians but plan to maintain speed or accelerate (NLI-C = 0.995); both hierarchical variants resolve the contradiction. Additional examples, including failure cases, appear in Appendix~\ref{sec:appendix}.

\section{Conclusion}
\label{sec:conclusion}

\paragraph{Scope.}
This paper operates at the GVQA \emph{reasoning level}: the Planning stage produces natural-language answers about what action the ego vehicle should take, following the DriveLM decomposition~\cite{drivelm}, not trajectory waypoints or closed-loop control signals. The safety-relevant contradictions our NLI metric detects (e.g., perceiving a crossing pedestrian yet planning to accelerate, or perceiving a red light yet planning to maintain speed) are \emph{reasoning-level} inconsistencies. We do not claim that reducing reasoning-level contradiction entails downstream trajectory correctness; as Drive-KD~\cite{Lian2026DriveKDMD} and Drive-P2D~\cite{Tang2026DriveP2DAP} independently demonstrate, perception and decision-making are weakly coupled in current VLMs, which motivates explicit coordination mechanisms like ours rather than assuming alignment emerges automatically.
The relationship between reasoning-level coherence and textual planning \emph{accuracy} (whether coherent plans are also correct plans) is an open question. We view this paper's consistency protocol as the diagnostic precondition for that next study.

\paragraph{Limitations.}
All experiments are conducted on DriveLM-nuScenes, currently the only public dataset providing typed hierarchical Perception/Prediction/Planning graph annotations on multi-view driving scenes. Generalizability of both the consistency protocol and the gated projector mechanism to other benchmarks remains untested. DriveLMM-o1~\cite{DriveLMMo1}, which provides stepwise reasoning chains, and candidate benchmarks augmented with cross-stage QA alignment are natural targets for follow-up evaluation.

\paragraph{Future work.}
Applying the implicit framework to a domain-adapted base model would test whether NLI gains extend to surface-level metrics. Our experience with the explicit variant suggests a concrete recipe: domain pretraining on driving VQA data with a coordinate-aware loss applied to the Perception adapter would ground the upstream stage in spatial scene structure before sequential training begins. This should reduce mixed-language generation and restore structural consistency while preserving the planning-stage NLI gains. Beyond consistency, the natural next evaluation axis is \emph{textual planning accuracy}: whether a coherent plan (one that does not contradict the model's own perception) is also a \emph{correct} plan. This would connect reasoning-level coherence to decision quality, the central open question for driving VLMs as reliable decision-making ``brains''. More broadly, the projector mechanism generalizes to any multi-step reasoning pipeline where intermediate hidden states should condition downstream generation.

\bibliography{colm2026_conference}
\bibliographystyle{colm2026_conference}
\section*{LLM Usage Disclosure}
Sections of this paper were drafted and revised with the assistance of
a large language model (Claude, Anthropic). All technical content,
experimental design, results, and scientific claims are solely the
authors' own. The LLM was not used to generate data, produce plots,
evaluate results, or originate research ideas.
\appendix

\section{Extended Qualitative Examples}
\label{sec:appendix}
This appendix supplements Section~\ref{sec:qualitative} with four additional scenes (two per variant), covering both success and failure cases of hierarchical context passing. Scene 162 (main paper, Figure~\ref{fig:qualitative_combined}) demonstrated the common success pattern; the examples below illustrate the remaining cases.

\subsection{Explicit Variant: Success Case (Scene 48)}

Scene 48 illustrates the \textit{red light $\to$ maintain speed} contradiction type. The flat baseline explicitly perceives a red light and a pedestrian to the front right, yet plans to maintain speed (C=1). Under injection-chaining, the model's perception shifts to a pedestrian-heavy construction scene without a traffic light; the structural checker finds no contradiction-triggering element (C=0).

\begin{figure}[h]
\centering
\includegraphics[width=0.9\columnwidth]{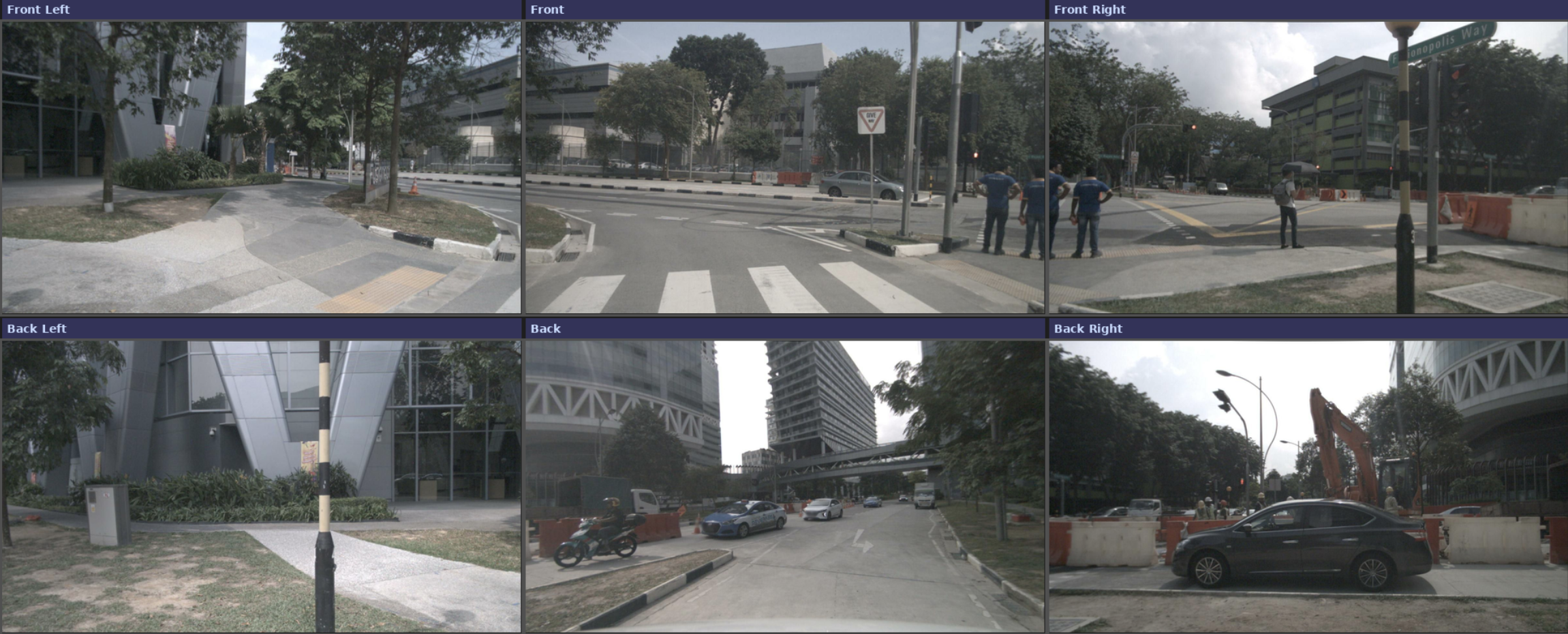}
\caption{Surround-view input for Scene 48 (DriveLM-nuScenes). Explicit variant success case.}
\label{fig:app_scene_48}
\end{figure}

\begin{table}[h]
\scriptsize
\setlength{\tabcolsep}{3pt}
\centering
\caption{Scene 48 -- Explicit variant success case. Injection-chain removes the red-light contradiction. \textbf{C} = structural contradiction count.}
\label{tab:app_explicit_best}
\begin{tabular}{@{}c|p{0.85\columnwidth}|c@{}}
\toprule
& \textbf{Stage Outputs} & \textbf{C} \\
\midrule
Flat &
  \textbf{P:} [\ldots] a pedestrian to the front right of the ego car, and a \underline{red light} to the front right of the ego car. \newline
  \textbf{Pr:} Keep going straight, turn left, turn right. \newline
  \textbf{Pl:} The action is to \underline{keep going at the same speed}. Turning left would be dangerous.
& \textbf{1} \\[3pt]
Inj. &
  \textbf{P:} There are many pedestrians, three barriers, two traffic cones [\ldots] The pedestrians are standing at the roadside. \newline
  \textbf{Pr:} One of the pedestrians will turn right. This will bring them into the ego vehicle's path. \newline
  \textbf{Pl:} The action is to \underline{keep going at the same speed}.
& 0 \\
\bottomrule
\end{tabular}
\end{table}

\subsection{Explicit Variant: Failure Case (Scene 489)}

Scene 489 is a case where the flat baseline is consistent but the injection-chain introduces a contradiction. The flat condition perceives both a pedestrian crosswalk and a green light; its plan avoids the pedestrian by turning right (C=0). Under injection-chaining, perception foregrounds the pedestrian crossing while the plan defaults to maintaining speed, triggering the \textit{pedestrian crossing $\to$ maintain speed} contradiction (NLI-C = 0.982 at the planning transition).

\begin{figure}[h]
\centering
\includegraphics[width=0.9\columnwidth]{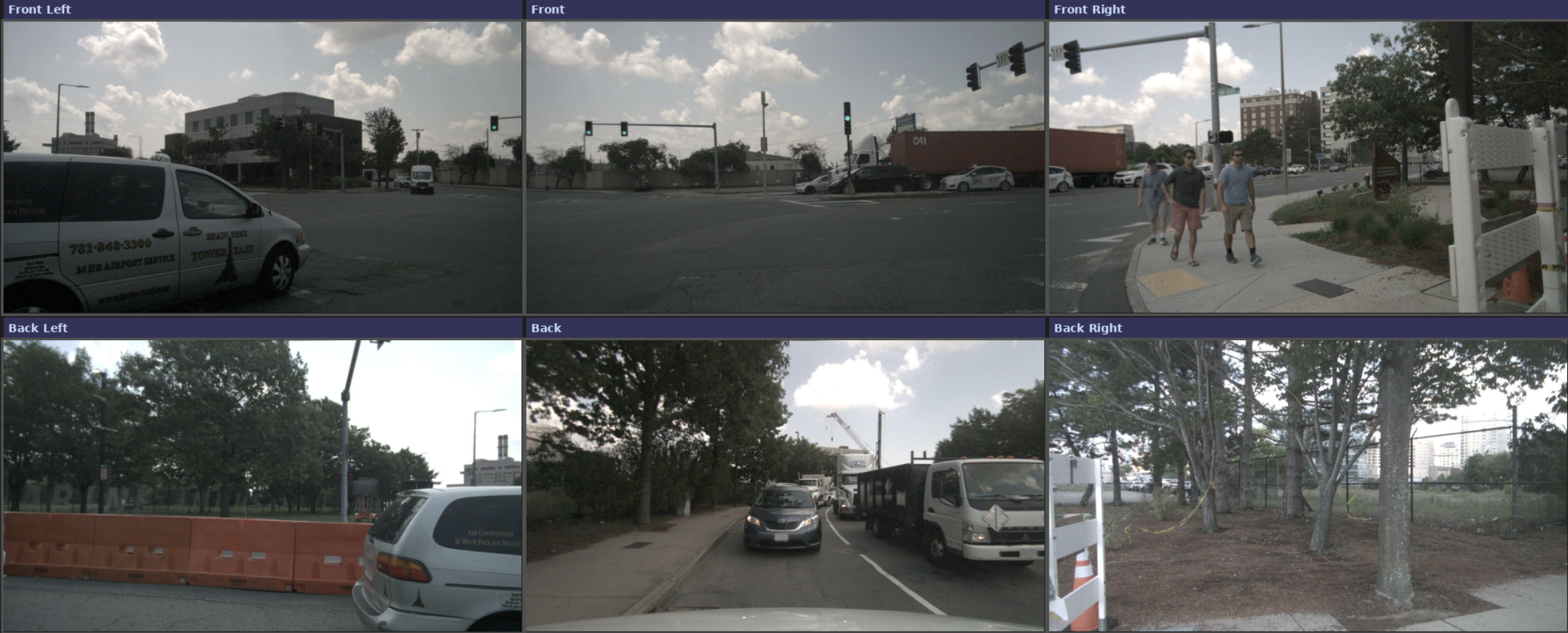}
\caption{Surround-view input for Scene 489 (DriveLM-nuScenes). Explicit variant failure case.}
\label{fig:app_scene_489}
\end{figure}

\begin{table}[h]
\scriptsize
\setlength{\tabcolsep}{3pt}
\centering
\caption{Scene 489 -- Explicit variant failure case. Injection-chain introduces a pedestrian-crossing contradiction. \textbf{C} = structural contradiction count.}
\label{tab:app_explicit_worst}
\begin{tabular}{@{}c|p{0.85\columnwidth}|c@{}}
\toprule
& \textbf{Stage Outputs} & \textbf{C} \\
\midrule
Flat &
  \textbf{P:} [\ldots] a black and white van, a black car, people walking on the sidewalk, a \underline{pedestrian crosswalk} [\ldots] and a \underline{green light}. [\ldots] \newline
  \textbf{Pr:} The brown car will turn right, and the blue sedan will continue going straight. They will be in the ego vehicle's path. \newline
  \textbf{Pl:} \underline{Turning right} is a good idea. Turning left is dangerous.
& 0 \\[3pt]
Inj. &
  \textbf{P:} A white sedan to the front left, a \underline{pedestrian crossing the road} to the front right, and a black sedan to the back. \newline
  \textbf{Pr:} The white sedan will change to the left lane, the pedestrian will continue moving forward, and the black sedan will change to the right lane. \newline
  \textbf{Pl:} The action of \underline{keep going at the same speed} is safe, the action of changing to the top right lane is dangerous.
& \textbf{1} \\
\bottomrule
\end{tabular}
\end{table}

\subsection{Implicit Variant: Success Case (Scene 120)}

Scene 120 demonstrates the \textit{green light $\to$ stop} contradiction type in the flat adapter baseline. The flat adapter perceives a green traffic light yet includes ``decelerate to a stop'' among safe actions, yielding one structural contradiction. The sequential PPP variant perceives the same green light and plans to maintain speed with no stopping action, producing a structurally consistent output (struct = 1.0, NLI-C = 0.001).

\begin{figure}[h]
\centering
\includegraphics[width=0.9\columnwidth]{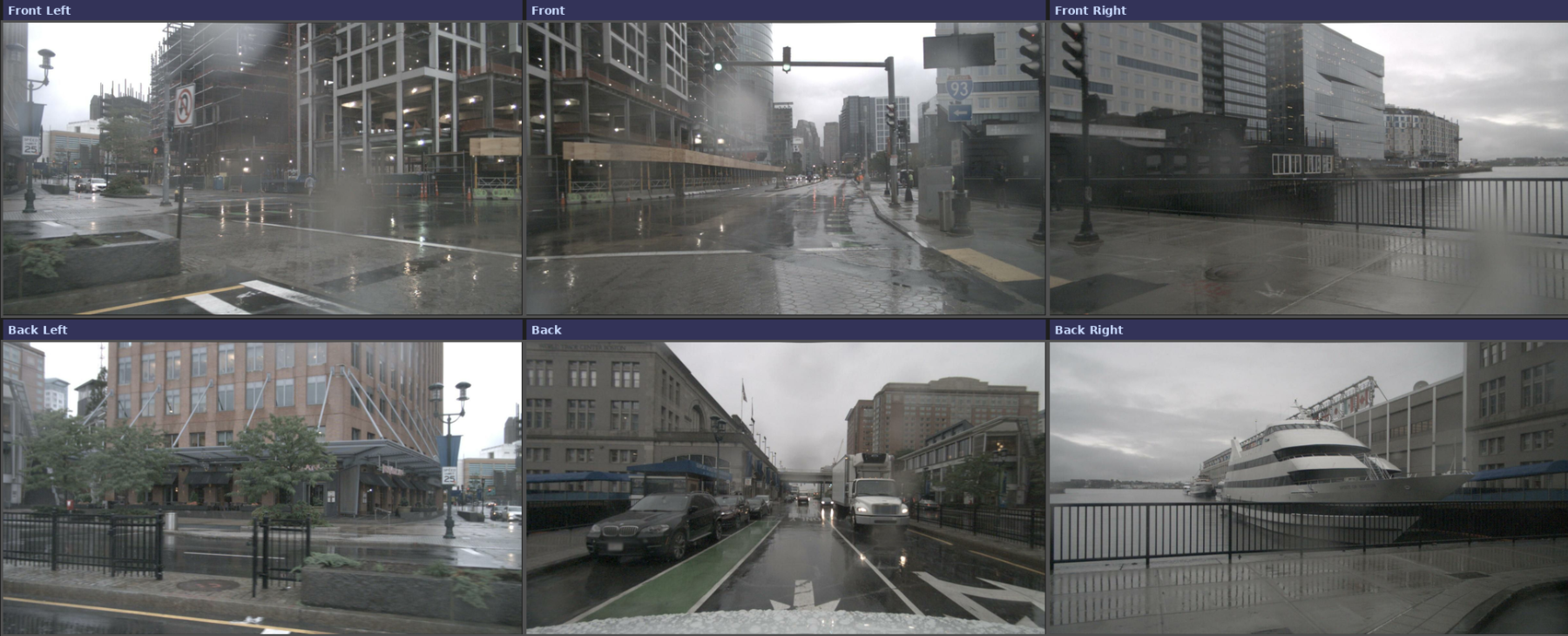}
\caption{Surround-view input for Scene 120 (DriveLM-nuScenes). Implicit variant success case.}
\label{fig:app_scene_120}
\end{figure}

\begin{table}[h]
\scriptsize
\setlength{\tabcolsep}{3pt}
\centering
\caption{Scene 120 -- Implicit variant success case. Sequential PPP removes the green-light/stop contradiction. \textbf{C} = structural contradiction count.}
\label{tab:app_implicit_best}
\begin{tabular}{@{}c|p{0.85\columnwidth}|c@{}}
\toprule
& \textbf{Stage Outputs} & \textbf{C} \\
\midrule
Flat &
  \textbf{P:} [\ldots] Traffic Light: \underline{Green light} visible. Road Markings: Visible on the wet road surface. [\ldots] \newline
  \textbf{Pr:} The traffic light is green, so the ego vehicle can proceed. The truck is stationary. There are no other vehicles in the immediate vicinity. \newline
  \textbf{Pl:} Gently decelerate gradually to a \underline{stop}, keep going at the same speed, and slightly offset to the left are safe actions. Turning left, turning right, and braking suddenly are dangerous actions.
& \textbf{1} \\[3pt]
Seq. &
  \textbf{P:} [\ldots] Traffic Light: \underline{Green light} visible. Road Markings: Crosswalk and lane markings on the wet road. Buildings under construction [\ldots] \newline
  \textbf{Pr:} None, no safety issue. \newline
  \textbf{Pl:} The action is to \underline{keep going at the same speed}. The reason is that there is no safety issue.
& 0 \\
\bottomrule
\end{tabular}
\end{table}

\subsection{Implicit Variant: Failure Case (Scene 360)}

Scene 360 shows the opposite pattern: the flat adapter baseline scores C=0 while the sequential PPP skip variant scores C=1. The flat plan hedges by listing both ``brake to a stop'' and ``keep going at the same speed''; the checker assigns C=0 because it finds a stopping action, even though the plan is internally contradictory. Under sequential skip, the prediction dismisses the pedestrian (``Neither of these objects will be in the ego vehicle's path''), and the plan commits to maintaining speed (C=1). The sequential plan is more decisive but penalized by the checker, illustrating a limitation of the rule-based metric: hedging with contradictory actions can score better than a single coherent action that omits a safety-relevant response.

\begin{figure}[h]
\centering
\includegraphics[width=0.9\columnwidth]{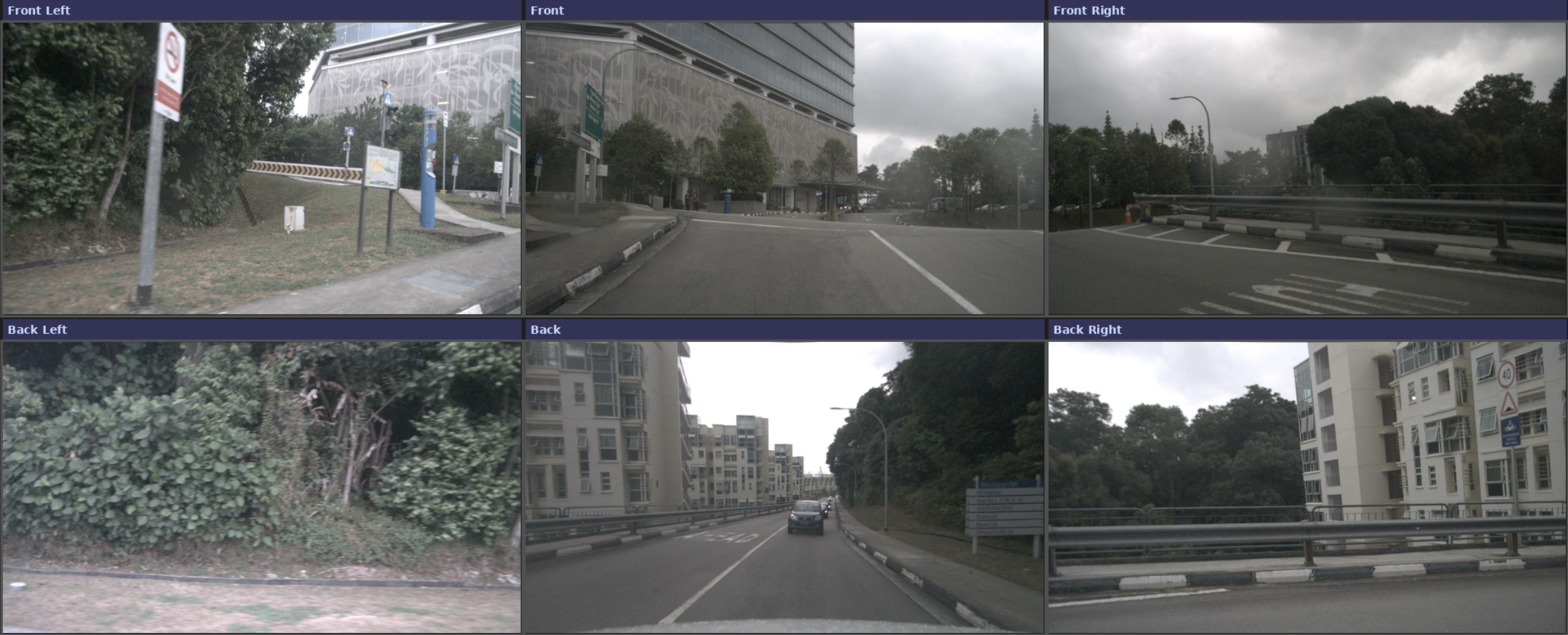}
\caption{Surround-view input for Scene 360 (DriveLM-nuScenes). Implicit variant failure case.}
\label{fig:app_scene_360}
\end{figure}

\begin{table}[h]
\scriptsize
\setlength{\tabcolsep}{3pt}
\centering
\caption{Scene 360 -- Implicit variant failure case. Sequential PPP skip introduces a pedestrian-crossing contradiction. \textbf{C} = structural contradiction count.}
\label{tab:app_implicit_worst}
\begin{tabular}{@{}c|p{0.85\columnwidth}|c@{}}
\toprule
& \textbf{Stage Outputs} & \textbf{C} \\
\midrule
Flat &
  \textbf{P:} [\ldots] \underline{Pedestrian}: Walking on the sidewalk [\ldots] Appears to be crossing the road. [\ldots] \newline
  \textbf{Pr:} [\ldots] Pedestrian crossing the road: may continue to cross, potentially entering the ego vehicle's path. The ego vehicle should be prepared to \underline{stop or slow down} [\ldots] \newline
  \textbf{Pl:} Gently \underline{brake to a stop}; keep going at the same speed.
& 0 \\[3pt]
Seq. &
  \textbf{P:} [\ldots] \underline{Pedestrian}: Walking on the sidewalk [\ldots] Appears to be crossing the road. [\ldots] \newline
  \textbf{Pr:} The black car in front will continue forward. The pedestrian will continue walking along the sidewalk. \underline{Neither of these objects will be in the ego vehicle's path.} \newline
  \textbf{Pl:} The action is to \underline{keep going at the same speed}. It is safe, and it is not dangerous.
& \textbf{1} \\
\bottomrule
\end{tabular}
\end{table}

\subsection{Summary of Appendix Examples}
The two dominant failure modes across all examples are: (1) \textit{pedestrian crossing $\to$ accelerate/maintain}, which accounts for the majority of structural contradictions in both variants (Section~\ref{sec:consistency}), and (2) \textit{traffic light $\to$ contradictory action} (red light with maintain speed, or green light with stop). In the success cases, hierarchical context consistently resolves these contradictions. In the failure cases, the hierarchical condition either re-introduces the pedestrian-crossing bias (explicit) or suppresses safety-relevant context through the learned projector (implicit). These patterns are consistent with the aggregate quantitative findings in Tables~\ref{tab:consistency} and~\ref{tab:answer_length}.
\begin{table}[t]
\small
\centering
\caption{Summary of qualitative examples. Scene 162 appears in the main paper (Figure~\ref{fig:qualitative_combined}); Scenes 48, 489, 120, 360 are detailed above. All outputs are in English.}
\label{tab:app_summary}
\begin{tabular}{llcccl}
\toprule
\textbf{Variant} & \textbf{Scene} & \textbf{Flat C} & \textbf{Hier.\ C} & \textbf{Type} & \textbf{Contradiction pattern} \\
\midrule
Explicit & 162 (main) & 1 & 0 & Best & ped crossing $\to$ maintain \\
Explicit & 48          & 1 & 0 & Best & red light $\to$ maintain \\
Explicit & 489         & 0 & 1 & Worst & ped crossing $\to$ maintain \\
\midrule
Implicit & 162 (main) & 1 & 0 & Best & ped crossing $\to$ accelerate \\
Implicit & 120         & 1 & 0 & Best & green light $\to$ stop \\
Implicit & 360         & 0 & 1 & Worst & ped crossing $\to$ maintain \\
\bottomrule
\end{tabular}
\end{table}

\end{document}